\documentclass{article}
\usepackage{dingbat}
\usepackage{spconf,amsmath,graphicx}
\usepackage{xcolor}
\usepackage{hyperref}

\title{SPACE: A Simulator for Physical  Interactions and Causal Learning in 3D Environments}
\name{Jiafei Duan$^{\star}$ \qquad Samson Yu Bai Jian$^{\ast}$ \qquad Cheston Tan$^{\dagger}$}

\address{duan0038@e.ntu.edu.sg, samson$\_$yu@mymail.sutd.edu.sg,  cheston-tan@i2r.a-star.edu.sg \\
$^{\star}$ Nanyang Technological University, Singapore \\
$^{\ast}$ Singapore University of Technology and Design\\
  $^{\dagger}$Agency for Science, Technology and Research (A*STAR)}

\begin{document}
%
\maketitle
\begin{abstract}
Recent advancements in deep learning, computer
vision, and embodied AI have given rise to synthetic causal
reasoning video datasets. These datasets facilitate the development of AI algorithms that can reason about physical interactions between objects. However, datasets thus far have primarily focused on elementary
physical events such as rolling or falling. There is currently
a scarcity of datasets that focus on the physical interactions that
humans perform daily with objects in the
real world. To address this scarcity, we introduce SPACE: A
Simulator for Physical Interactions and Causal Learning in
3D Environments. The SPACE simulator allows us to generate the SPACE dataset, a synthetic video dataset in a 3D environment, to systematically evaluate physics-based models on a range of physical causal reasoning tasks. Inspired by daily object interactions, the SPACE dataset comprises videos depicting three types of physical events: containment, stability and contact. These events make up the vast majority of the basic physical interactions between objects. We then further evaluate it with a state-of-the-art physics-based deep model and show that the SPACE dataset improves the learning of intuitive physics with an approach inspired by curriculum learning. \textcolor{blue}{Repository: \url{https://github.com/jiafei1224/SPACE}}
\end{abstract}
\begin{keywords}
3D Simulators, Physical interaction, Causal Learning, Computer Vision
\end{keywords}
\section{Introduction}
Traditionally, it is believed that infants have very little understanding of the physical world due to their poor performance on object-manipulation tasks \cite{piaget1954construction,piaget1952origins}. However, with further research conducted via new visual-attention methods, it has been revealed that, contrary to traditional beliefs, even young infants possess a wealth of knowledge about the physical world and have the innate ability to perform causal reasoning over physical interactions between objects \cite{baillargeon1995model,hespos2001infants,needham1997object,sperber1995causal}. In addition, young infants develop their sensitivity to kinetic information for depth in the very early stages of development, which allows them to foster a response for expansion and contraction patterns specifying approach to three-dimensional objects \cite{kayed2000timing,schmuckler1998looming,shirai2004asymmetry}. Beyond that, infants can also learn to recognise the objects from those interactions when they are as young as nine months old \cite{shinskey2014picturing}. Thus, as a result of all these attributes, infants can observe a wide range of physical interactions (e.g. containment, stability and contact) between objects of various geometric shapes and further develop a causal understanding for fundamental physical interactions in their preoperational stage of development \cite{huitt2003piaget}. Furthermore, according to the theory of accommodation \cite{piaget1976piaget}, infants can further modify their basic understanding of physical interactions into higher order tasks which incorporate a mixture of those fundamental physical interactions. Hence, it is of paramount importance for infants to have fully developed their innate ability in understanding those fundamental physical interactions before moving on to the higher order tasks. Equivalently, it is vital for AI models that aim to improve their understanding and reasoning capabilities of real-world object interactions, to be first trained on those fundamental physical interactions. 

\begin{figure}[t]
\begin{center}
\includegraphics[width=\linewidth]{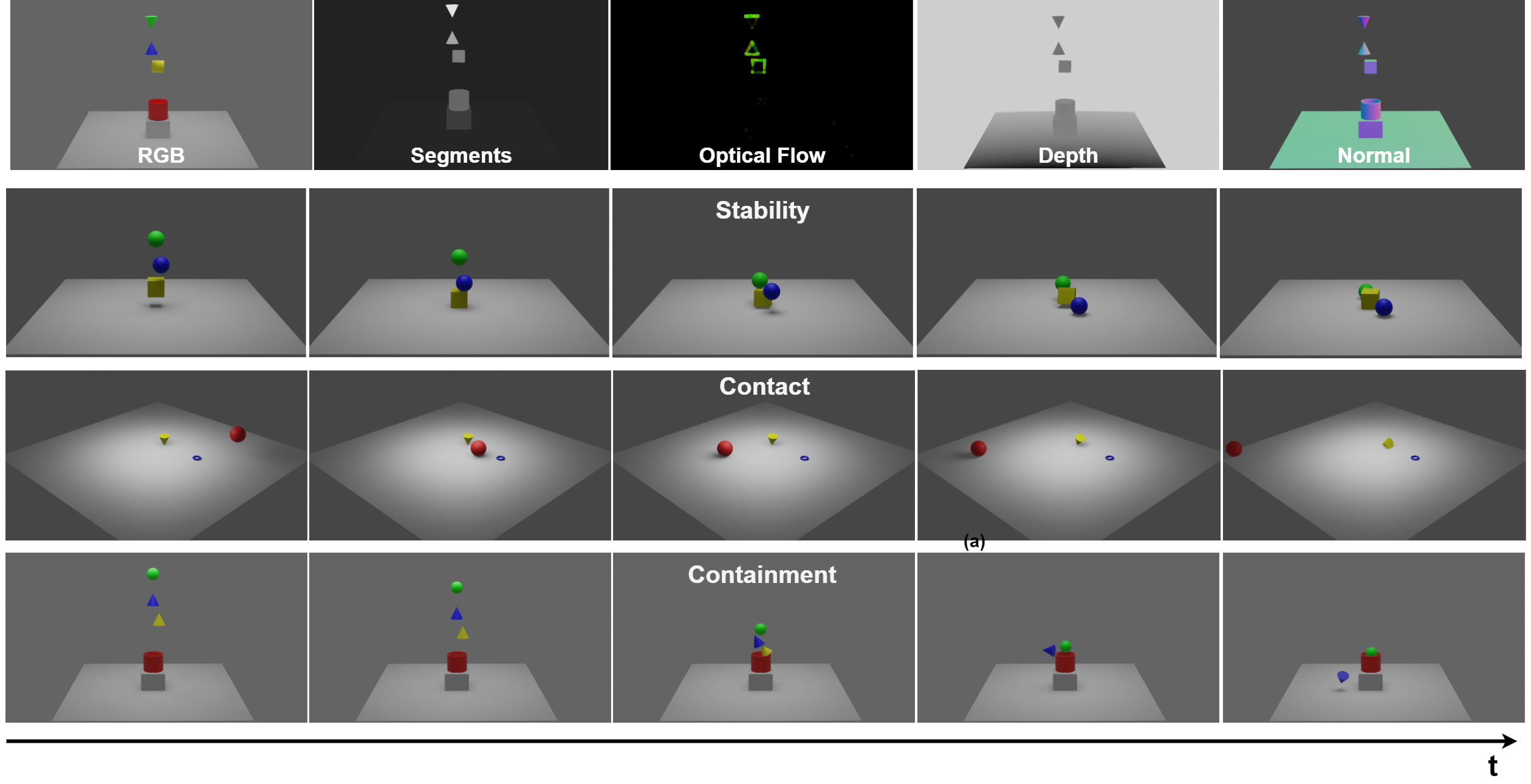}
\end{center}
  \caption{ Top row: visual data attributes for one example frame comprises of RGB, object segmentation, optical flow, depth, and surface normal vector. Bottom three rows: example frames from the three physical interactions.}
\label{fig:long3}
\end{figure}

This paper aims to study the significance of causal reasoning in fundamental physical interactions between objects in a 3D environment. Our work draws inspiration from a few recent 3D visual reasoning datasets \cite{baradel2019cophy,girdhar2019cater,yi2019clevrer} in terms of their methodologies for synthesizing large-scale physical interaction datasets in a 3D environment. As such, the paper composed of two parts: 1) the SPACE simulator for synthesizing of various physical interactions between objects in 3D environment with a large degree of conditional variances and accurate physics-based interactions, and 2) the SPACE dataset, a large-scale video dataset with over 15k unique videos and over 2 million frames of the three fundamental physical interactions: containment, stability and contact, as shown in Figure \ref{fig:long3}. Furthermore, we evaluate the SPACE dataset on PhyDNet \cite{leguen20phydnet}, a state-of-the-art model for physics-based future frame prediction and show that by pre-training on our SPACE dataset, we can improve the physics-based model's performance on higher order real-world human action datasets after fine-tuning in a way similar to curriculum learning \cite{10.1145/1553374.1553380}.

\section{Related Work} 
The notion of drawing inspiration from conventional concepts in neuroscience and psychology to understand human perception of physical dynamics and causal reasoning has been around for some time \cite{aggarwal2011human,cao2020long,kong2018human,liang2019peeking}. However, those works primarily focused on human intuitions of recognizing or predicting motions. But with the advancement and rise of deep learning, computer graphics, and embodied AI \cite{duan2021survey,kobbelt2004survey,lecun2015deep}, there has been a paradigm shift towards generating synthetic datasets that range from simple 2D cartoons \cite{gordon2016commonsense,netanyahu2021phase,zitnick2014adopting} to realistic interaction in 3D environments \cite{baradel2019cophy,duan2020actionet,girdhar2019cater,johnson2017clevr,shu2021agent,yi2019clevrer}, all with the aim to explore machine perception of physics and causal reasoning on a deeper level. However, only datasets from CLEVRER \cite{yi2019clevrer}, CoPhy \cite{baradel2019cophy}, and CATER \cite{girdhar2019cater} are most relevant to our work.

\begin{figure}[t]
\begin{center}
  \includegraphics[width=\linewidth]{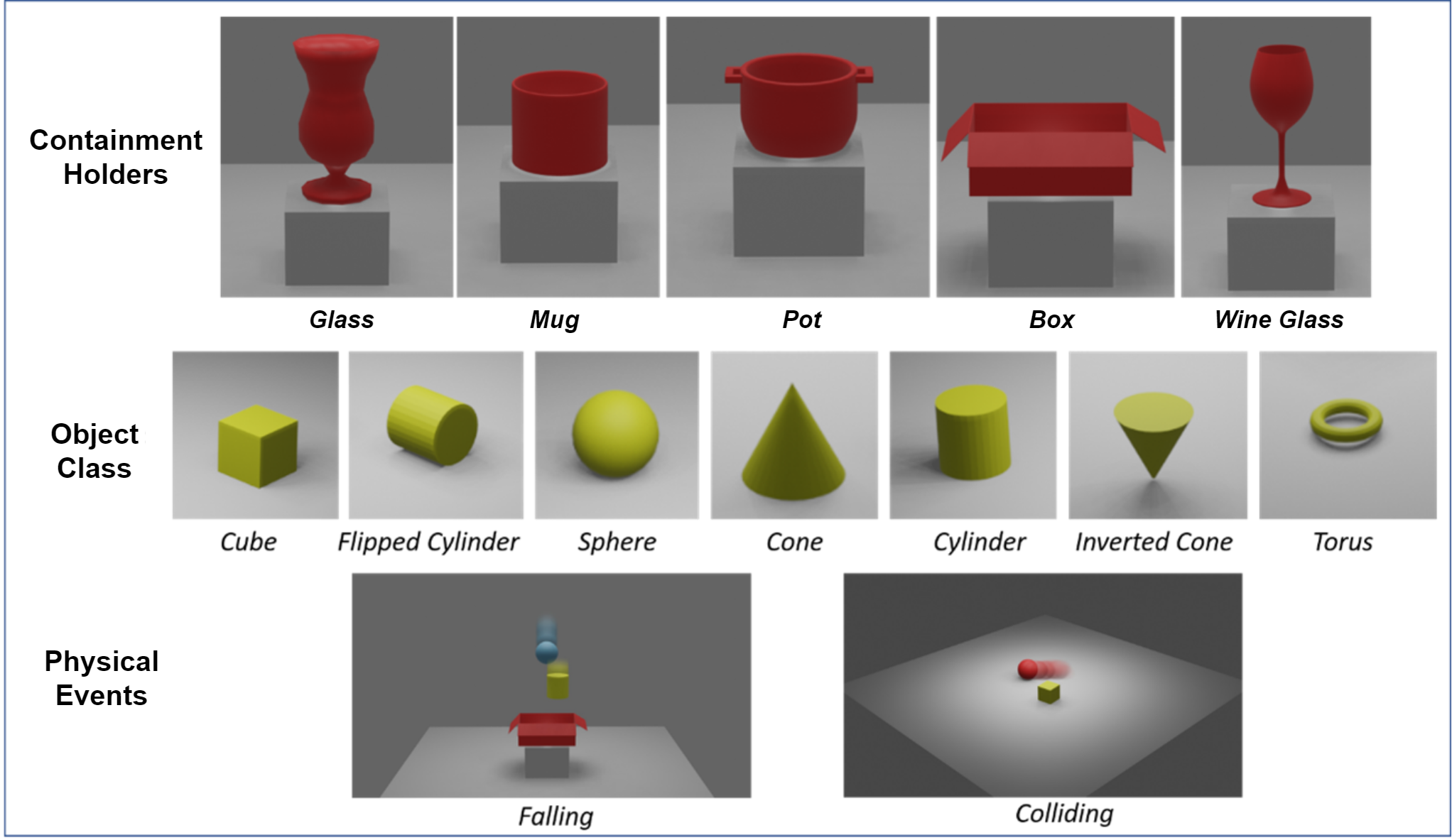}
\end{center}
  \caption{Environment setup, and all the available interactable objects and containment holders (for containment task only) spawned for the two physical events.}
\label{fig:long2}
\end{figure}
CLEVRER \cite{yi2019clevrer} is a collision-based video dataset comprising over 20,000 synthetic videos of object collision and 300,000 questions and answers on four newly proposed categories of questions: descriptive, explanatory, predictive, and counterfactual. A Neuro-Symbolic Dynamic Reasoning model was further proposed to tackles the dataset.

CoPhy \cite{baradel2019cophy} developed a model for causal physical reasoning in a synthetic 3D environment. However, they propose a model that learns the physical dynamics from a counterfactual setting and based on the given scenarios to make the next frame prediction. The CoPhy framework allows for the generation of 300k different scenarios in a 3D environment, such as a tower of blocks falling or objects colliding. 


CATER \cite{girdhar2019cater} built a video dataset with observable and controllable objects in a 3D environment and requires the model to have spatiotemporal understanding to solve it. It consists of three different tasks: atomic action recognition, compositional action recognition, and snitch localization (an object permanence task where the goal is to classify the final location of the snitch object within a 2D grid space), with 5,500 videos for each of the tasks. 


Even though our SPACE dataset might share similar visual modalities as the three datasets, but the nature of its physical interactions and the complexity in dynamics are very different. The purpose of the SPACE dataset is to serve as the building blocks for curriculum learning \cite{10.1145/1553374.1553380} in more complex real-world object interaction tasks. Hence, our SPACE dataset is rich in complexity and diversity, making it a suitable dataset for training AI models to learn the intuitive physics behind object interactions.

\section{SPACE Dataset}
\subsection{Overview}
Figure \ref{fig:long3} shows the composition of the SPACE dataset, which is made up of three novel video datasets that are generated from the scenarios synthesized based on three fundamental physical interactions: \emph{containment}, \emph{stability}, and \emph{contact} in a 3D environment. Each of the physical interaction scenarios is synthesized by our SPACE simulator, which is developed using Blender \cite{blender}, an open-source 3D computer graphics software with a Python API. The SPACE simulator allows users to generate various scenarios for each of the physical interactions with different cinematic conditions, object classes and position settings as shown in Figure \ref{fig:long2}. Each physical interaction scenario generated has several parameters such as object classes $O$, number of objects spawned $N$, spawn locations $L$, and containment holders $C$ (only applicable to the containment task). These parameters are selected because they affect the physical dynamics significantly, thus helping to create more diverse set of scenarios for each category of the physical interactions.

\textbf{Containment.} This physical interaction is demonstrated by synthesizing up to three objects of various object classes to be spawned at different locations with predefined heights which are directly above one of the containment holders as illustrated in Figure \ref{fig:long3}. The data will capture the final coordinates of each object to determine if it is contained within the containment holder $label_{containment}=\{0,1\}$.


\textbf{Stability.} This physical interaction is demonstrated via having up to three objects of various object classes spawning at different locations with predefined heights which are directly above the origin of the plane as shown in Figure \ref{fig:long3}. The stability $label_{stability}=\{0,1\}$ of each object can be determined through the differences in the angular rotational motion of each object measured throughout the scene.  


\textbf{Contact.} This physical interaction is demonstrated via having up to three objects of various object classes spawned at different locations and at a ground level height as shown in Figure \ref{fig:long3}. After the scene has been initialized, a ball of fixed mass and shape will be spawned and moved across the plane at a speed of $1m/s$ with a fixed trajectory. Whether each object has been contacted $label_{contact}=\{0,1\}$ are determined by the changes in its initial and final positions.

\subsection{Procedural Generation}
To synthesize all the physical interaction scenarios, the SPACE simulator will sample from a range of one to three objects $N=\{1,2,3\}$ to be spawned in the scene, and the class of each object are then sampled from $O=$ \{\emph{cylinder, cone, inverted cone, cube, torus, sphere, flipped cylinder}\}.

However, there are differences in certain settings between each category of the physical interactions. For containment, the object(s) will be spawned directly above the containment holder with height differences of $\pm0.5$ between two consecutive objects. The containment holder is sampled from $C=$ \{\emph{wine glass, glass, mug, pot, box}\}.
For stability, the object(s) will be spawned directly above the origin $(0,0,z)$ with its \emph{x} and \emph{y} coordinates sampled from $\{0.3,0.2,0.1,0,-0.1,-0.2,-0.3\}$.
For contact, the object(s) will be spawned within a $6\times6$ grid with an incremental or decremental value of 1 for both the \emph{x} and \emph{y} axes at ground level.

\begin{figure}[t]
\begin{center}
  \includegraphics[width=\linewidth]{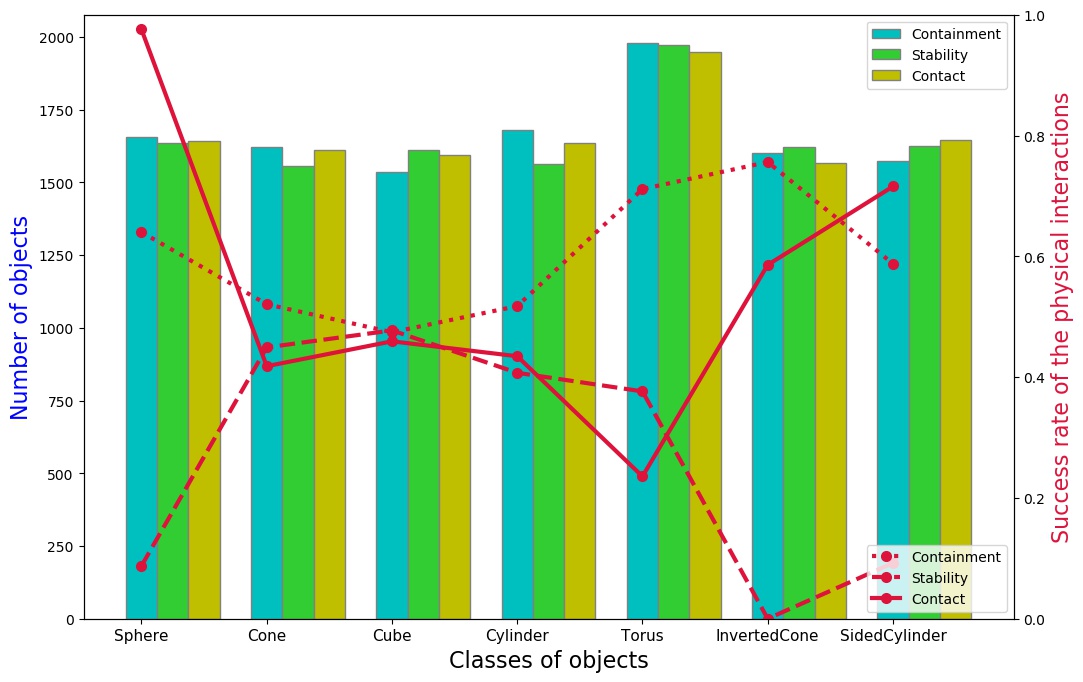}
\end{center}
  \caption{Dataset analysis for number of objects and success rate of physical interactions by object classes.}
\label{fig:long}
\end{figure}


\subsection{Dataset Structure}
 There are 15,000 unique scene instances generated for the SPACE simulator, with 5,000 scenes for each category of physical interactions. From there, we collect 15,000 videos lasting 3 seconds with a frame rate of 50 frames per second (FPS), which total up to 2,250,000 frames. Besides the RGB frames, we also provide the segmentation map, optical flow map, depth map and surface normal vector map as shown in Figure \ref{fig:long3}. We further perform an in-depth analysis of the composition of each geometric object within each category of physical interactions and analyze the success rate for each of the geometric objects within the scene. The success rate for each of the geometric objects within their physical interaction task is a clear indication of how the geometric shape of the object might have on the overall outcome of physical interactions, as seen in Figure \ref{fig:long}.

\section{Experiments}
In our experiments, we aim to show that the SPACE dataset is useful in helping physics-based deep models learn robust physical dynamics for downstream tasks.

\subsection{Task \& Dataset}
In this paper, we illustrate the efficacy of our dataset for improving physics-based models by comparing two different experimental setups on the future frame prediction task \cite{liu2018future}. For this task, we make use of our SPACE dataset and the UCF101 dataset \cite{soomro2012ucf101}. The UCF101 dataset consists of realistic action videos, collected from YouTube, with 101 action categories.

Specifically, we want to show that a transfer learning setup that leverages on pre-training with the SPACE dataset outperforms one that is limited to pre-training on the UCF101 dataset, when both models are subsequently fine-tuned and evaluated on the UCF101 dataset. Our setup is also inspired by curriculum learning \cite{10.1145/1553374.1553380}, where data is presented to the model in a way that gradually illustrates more complex data. In this case, UCF101 represents the increased complexity, since it is situated in the real world while our SPACE dataset is in a synthetic environment.

\subsection{Network Architecture}
We select a state-of-the-art physics-based model, PhyDNet \cite{leguen20phydnet} for our experiments. PhyDNet introduces a two-branch deep architecture that disentangles the learning of physical dynamics and unknown complementary information that do not correspond to any prior model. In PhyDNet, physics knowledge is represented by the learning of partial differential equations (PDE) and is used to enforce physical constraints for future frame prediction. Unlike previous models, PhyDNet allows for a generic class of linear PDEs through the varying of differential orders, enabling it to account for more classical models (e.g. the heat equation, the wave equation and etc). To learn the unknown complementary information, PhyDNet uses a generic ConvLSTM \cite{xingjian2015convolutional}.

\subsection{Experimental Settings}
\label{sec:experimental_settings}
For all experiments, we use a batch size of 2 and a sequence length of 50 frames. We remove UCF101 samples that have less than 50 frames. Within the 50 frames, the first 5 frames are set as the input sequence, and the remaining 45 are to be generated. We train for 20 and 30 epochs for pre-training and fine-tuning respectively. We use an adaptive learning rate that follows the original PhyDNet setup. During fine-tuning for both SPACE and UCF101, we set the starting learning rate to be the final learning rate in the UCF101 pre-training, and maintain the adaptive learning rate. Like in the original PhyDNet setup, we also employ teacher forcing \cite{williams1989learning}, we set the reduction rate in teacher forcing probability threshold to be the same as that of PhyDNet with each epoch. We reset this threshold at the start of the fine-tuning process with the same reduction rate.

For SPACE, we sample 50 frames by taking 1 frame every 3 frames, since each scene has a sequence of 150 frames. We take 1,000 instances for each task in SPACE: containment, stability and contact for a total of 3,000 instances. For each task, we take 200 samples as validation data in a 8:2 ratio. Likewise, we split UCF101 into training and validation data by categories in a 8:2 ratio.

We vary the dataset composition in our experiments for both SPACE and UCF101 to test the importance of the three SPACE tasks in helping physics-based deep models learn robust physical dynamics. Specifically, we have two scenarios:
1) 3 UCF101 categories are filmed with a fixed camera, and encapsulate physical dynamics from at least two of the three SPACE tasks (UCF3), while having minimal human intervention/actions throughout the video,
and 2) individual SPACE tasks evaluated on scenario 2 during fine-tuning.

\subsection{Evaluation Metric}
We evaluate our experimental setups using an evaluation metric commonly used in video prediction: Mean Squared Error (MSE). This metric computes the quality of generated frames in comparison to their respective ground-truth frames. The MSE is averaged for each frame of the output sequence. Lower MSE indicates better performance.
\begin{figure}[t]
\begin{center}
  \includegraphics[width=\linewidth]{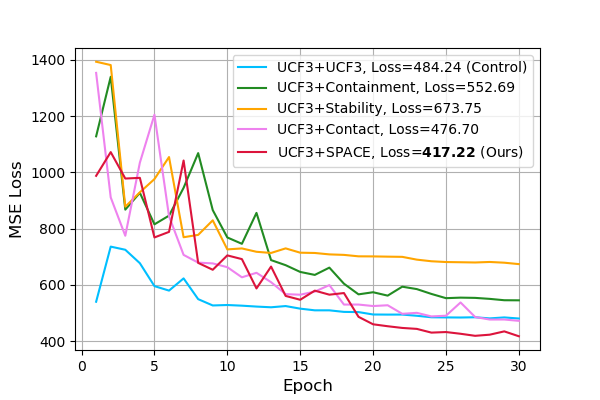}
\end{center}
  \caption{Experimental results for scenarios in Section \ref{sec:experimental_settings}. For scenario 1, we compare the effects of UCF3 and SPACE pre-training (cyan and red lines), and for scenario 2, we compare the effects of individual SPACE tasks (green, orange and pink lines).}
\label{fig:result}
\end{figure}
\subsection{Results}
Our results show that the SPACE dataset helps physics-based model, PhyDNet to learn robust physical dynamics. In Figure \ref{fig:result}, we show that the validation loss in experimental scenario 1 (where we train the models on UCF3) is lower in the case where the model is pre-trained on the SPACE dataset. We also show that each of the three fundamental physical interactions in the SPACE dataset is essential for the learning of robust physical dynamics, since the experimental results for scenario 2 are worse than when they are combined in scenario 1. Furthermore, experimental scenario 1 with SPACE pre-training has the overall lowest validation MSE loss after 30 epochs of fine-tuning on UCF3.

\subsection{Conclusion}
Children learn the causality and intuitive physics behind object interaction by first learning about fundamental physical interaction concepts such as contact, containment and stability. Similarly, for AI models to improve their understanding of object interactions, they should also be trained in an approach like curriculum learning. In this work, we propose SPACE, a simulator for physical interactions and causal learning in 3D environments, that composes a large-scale synthetic video dataset for three fundamental physical interaction tasks. Our experiments show that the pre-training done the on SPACE dataset improves a state-of-the-art physics-based deep learning model's performance on real-world object interaction tasks.

\section*{Acknowledgments}
This research is supported by the Agency for Science, Technology and Research (A*STAR), Singapore under its AME
Programmatic Funding Scheme (Award \#A18A2b0046) and the National Research Foundation, Singapore under its NRF-ISF Joint Call (Award NRF2015-NRF-ISF001-2541). We would also like to thank the Machine Learning and Data Analytics Lab at School of Electrical and Electronic Engineering, Nanyang Technological University and DeCLaRe Lab, Singapore University of Technology and Design for the computational resources used for this work.

\bibliographystyle{IEEEbib}
\bibliography{strings,refs}

\begin{thebibliography}{10}

\bibitem{piaget1954construction}
J~Piaget,
\newblock ``The construction of reality in the child (m. cook, trans.). new
  york, ny, us,''
\newblock {\em Basic Books. https://doi. org/10}, vol. 1037, pp. 11168--000,
  1954.

\bibitem{piaget1952origins}
Jean Piaget and Margaret~Trans Cook,
\newblock ``The origins of intelligence in children.,''
\newblock 1952.

\bibitem{baillargeon1995model}
Renee Baillargeon,
\newblock ``A model of physical reasoning in infancy,''
\newblock in {\em In C. Rovee-Collier, \& LP Lipsitt (Eds.), Advances in
  infancy research}. Citeseer, 1995.

\bibitem{hespos2001infants}
Susan~J Hespos and Ren{\'e}e Baillargeon,
\newblock ``Infants' knowledge about occlusion and containment events: A
  surprising discrepancy,''
\newblock {\em Psychological Science}, vol. 12, no. 2, pp. 141--147, 2001.

\bibitem{needham1997object}
Amy Needham, Ren{\'e}e Baillargeon, and Lisa Kaufman,
\newblock ``Object segregation in infancy.,''
\newblock {\em Advances in infancy research}, 1997.

\bibitem{sperber1995causal}
Dan Sperber, David Premack, Ann~James Premack, et~al.,
\newblock {\em Causal cognition: A multidisciplinary debate},
\newblock Number Sirsi) i9780198523147. Clarendon Press Oxford, 1995.

\bibitem{kayed2000timing}
Nanna~S{\o}nnichsen Kayed and Audrey van~der Meer,
\newblock ``Timing strategies used in defensive blinking to optical collisions
  in 5-to 7-month-old infants,''
\newblock {\em Infant Behavior and Development}, vol. 23, no. 3-4, pp.
  253--270, 2000.

\bibitem{schmuckler1998looming}
Mark~A Schmuckler and Nicole~S Li,
\newblock ``Looming responses to obstacles and apertures: The role of accretion
  and deletion of background texture,''
\newblock {\em Psychological Science}, vol. 9, no. 1, pp. 49--52, 1998.

\bibitem{shirai2004asymmetry}
Nobu Shirai, So~Kanazawa, and Masami~K Yamaguchi,
\newblock ``Asymmetry for the perception of expansion/contraction in infancy,''
\newblock {\em Infant Behavior and Development}, vol. 27, no. 3, pp. 315--322,
  2004.

\bibitem{shinskey2014picturing}
Jeanne~L Shinskey and Liza~J Jachens,
\newblock ``Picturing objects in infancy,''
\newblock {\em Child development}, vol. 85, no. 5, pp. 1813--1820, 2014.

\bibitem{huitt2003piaget}
William Huitt and John Hummel,
\newblock ``Piaget's theory of cognitive development,''
\newblock {\em Educational psychology interactive}, vol. 3, no. 2, pp. 1--5,
  2003.

\bibitem{piaget1976piaget}
Jean Piaget,
\newblock ``Piaget’s theory,''
\newblock in {\em Piaget and his school}, pp. 11--23. Springer, 1976.

\bibitem{baradel2019cophy}
Fabien Baradel, Natalia Neverova, Julien Mille, Greg Mori, and Christian Wolf,
\newblock ``Cophy: Counterfactual learning of physical dynamics,''
\newblock {\em arXiv preprint arXiv:1909.12000}, 2019.

\bibitem{girdhar2019cater}
Rohit Girdhar and Deva Ramanan,
\newblock ``Cater: A diagnostic dataset for compositional actions and temporal
  reasoning,''
\newblock {\em arXiv preprint arXiv:1910.04744}, 2019.

\bibitem{yi2019clevrer}
Kexin Yi, Chuang Gan, Yunzhu Li, Pushmeet Kohli, Jiajun Wu, Antonio Torralba,
  and Joshua~B Tenenbaum,
\newblock ``Clevrer: Collision events for video representation and reasoning,''
\newblock {\em arXiv preprint arXiv:1910.01442}, 2019.

\bibitem{leguen20phydnet}
Vincent Le~Guen and Nicolas Thome,
\newblock ``Disentangling physical dynamics from unknown factors for
  unsupervised video prediction,''
\newblock in {\em Computer Vision and Pattern Recognition (CVPR)}. 2020.

\bibitem{10.1145/1553374.1553380}
Yoshua Bengio, J\'{e}r\^{o}me Louradour, Ronan Collobert, and Jason Weston,
\newblock ``Curriculum learning,''
\newblock in {\em Proceedings of the 26th Annual International Conference on
  Machine Learning}, New York, NY, USA, 2009, ICML '09, p. 41–48, Association
  for Computing Machinery.

\bibitem{aggarwal2011human}
Jake~K Aggarwal and Michael~S Ryoo,
\newblock ``Human activity analysis: A review,''
\newblock {\em ACM Computing Surveys (CSUR)}, vol. 43, no. 3, pp. 1--43, 2011.

\bibitem{cao2020long}
Zhe Cao, Hang Gao, Karttikeya Mangalam, Qi-Zhi Cai, Minh Vo, and Jitendra
  Malik,
\newblock ``Long-term human motion prediction with scene context,''
\newblock in {\em European Conference on Computer Vision}. Springer, 2020, pp.
  387--404.

\bibitem{kong2018human}
Yu~Kong and Yun Fu,
\newblock ``Human action recognition and prediction: A survey,''
\newblock {\em arXiv preprint arXiv:1806.11230}, 2018.

\bibitem{liang2019peeking}
Junwei Liang, Lu~Jiang, Juan~Carlos Niebles, Alexander~G Hauptmann, and
  Li~Fei-Fei,
\newblock ``Peeking into the future: Predicting future person activities and
  locations in videos,''
\newblock in {\em Proceedings of the IEEE/CVF Conference on Computer Vision and
  Pattern Recognition}, 2019, pp. 5725--5734.

\bibitem{duan2021survey}
Jiafei Duan, Samson Yu, Hui~Li Tan, Hongyuan Zhu, and Cheston Tan,
\newblock ``A survey of embodied ai: From simulators to research tasks,''
\newblock {\em arXiv preprint arXiv:2103.04918}, 2021.

\bibitem{kobbelt2004survey}
Leif Kobbelt and Mario Botsch,
\newblock ``A survey of point-based techniques in computer graphics,''
\newblock {\em Computers \& Graphics}, vol. 28, no. 6, pp. 801--814, 2004.

\bibitem{lecun2015deep}
Yann LeCun, Yoshua Bengio, and Geoffrey Hinton,
\newblock ``Deep learning,''
\newblock {\em nature}, vol. 521, no. 7553, pp. 436--444, 2015.

\bibitem{gordon2016commonsense}
Andrew Gordon,
\newblock ``Commonsense interpretation of triangle behavior,''
\newblock in {\em Proceedings of the AAAI Conference on Artificial
  Intelligence}, 2016, vol.~30.

\bibitem{netanyahu2021phase}
Aviv Netanyahu, Tianmin Shu, Boris Katz, Andrei Barbu, and Joshua~B Tenenbaum,
\newblock ``Phase: Physically-grounded abstract social events for machine
  social perception,''
\newblock {\em arXiv preprint arXiv:2103.01933}, 2021.

\bibitem{zitnick2014adopting}
C~Lawrence Zitnick, Ramakrishna Vedantam, and Devi Parikh,
\newblock ``Adopting abstract images for semantic scene understanding,''
\newblock {\em IEEE transactions on pattern analysis and machine intelligence},
  vol. 38, no. 4, pp. 627--638, 2014.

\bibitem{duan2020actionet}
Jiafei Duan, Samson Yu, Hui~Li Tan, and Cheston Tan,
\newblock ``Actionet: An interactive end-to-end platform for task-based data
  collection and augmentation in 3d environment,''
\newblock in {\em 2020 IEEE International Conference on Image Processing
  (ICIP)}. IEEE, 2020, pp. 1566--1570.

\bibitem{johnson2017clevr}
Justin Johnson, Bharath Hariharan, Laurens Van Der~Maaten, Li~Fei-Fei,
  C~Lawrence~Zitnick, and Ross Girshick,
\newblock ``Clevr: A diagnostic dataset for compositional language and
  elementary visual reasoning,''
\newblock in {\em Proceedings of the IEEE Conference on Computer Vision and
  Pattern Recognition}, 2017, pp. 2901--2910.

\bibitem{shu2021agent}
Tianmin Shu, Abhishek Bhandwaldar, Chuang Gan, Kevin~A Smith, Shari Liu, Dan
  Gutfreund, Elizabeth Spelke, Joshua~B Tenenbaum, and Tomer~D Ullman,
\newblock ``Agent: A benchmark for core psychological reasoning,''
\newblock {\em arXiv preprint arXiv:2102.12321}, 2021.

\bibitem{blender}
Blender~Online Community,
\newblock {\em Blender - a 3D modelling and rendering package},
\newblock Blender Foundation, Stichting Blender Foundation, Amsterdam, 2018.

\bibitem{liu2018future}
Wen Liu, Weixin Luo, Dongze Lian, and Shenghua Gao,
\newblock ``Future frame prediction for anomaly detection--a new baseline,''
\newblock in {\em Proceedings of the IEEE conference on computer vision and
  pattern recognition}, 2018, pp. 6536--6545.

\bibitem{soomro2012ucf101}
Khurram Soomro, Amir~Roshan Zamir, and Mubarak Shah,
\newblock ``Ucf101: A dataset of 101 human actions classes from videos in the
  wild,''
\newblock {\em arXiv preprint arXiv:1212.0402}, 2012.

\bibitem{xingjian2015convolutional}
SHI Xingjian, Zhourong Chen, Hao Wang, Dit-Yan Yeung, Wai-Kin Wong, and
  Wang-chun Woo,
\newblock ``Convolutional lstm network: A machine learning approach for
  precipitation nowcasting,''
\newblock in {\em Advances in neural information processing systems}, 2015, pp.
  802--810.

\bibitem{williams1989learning}
Ronald~J Williams and David Zipser,
\newblock ``A learning algorithm for continually running fully recurrent neural
  networks,''
\newblock {\em Neural computation}, vol. 1, no. 2, pp. 270--280, 1989.

\end{thebibliography}

\end{document}